\documentclass[sigconf]{acmart}

\usepackage{amsthm}

\usepackage{float}
\newenvironment{myitemize}{\begin{list}{$\bullet$}
{\setlength{\topsep}{1mm}
\setlength{\itemsep}{0.25mm}
\setlength{\parsep}{0.25mm}
\setlength{\itemindent}{0mm}
\setlength{\partopsep}{0mm}
\setlength{\labelwidth}{15mm}
\setlength{\leftmargin}{4mm}}}{\end{list}}
\usepackage[ruled,linesnumbered]{algorithm2e}
\SetKwRepeat{Do}{do}{while}
\usepackage{multirow}

\usepackage{array}
\newcolumntype{L}[1]{>{\raggedright\let\newline\\\arraybackslash\hspace{0pt}}m{#1}}
\newcolumntype{C}[1]{>{\centering\let\newline\\\arraybackslash\hspace{0pt}}m{#1}}
\newcolumntype{R}[1]{>{\raggedleft\let\newline\\\arraybackslash\hspace{0pt}}m{#1}}

\usepackage{balance} 

\renewcommand\footnotetextcopyrightpermission[1]{}

\title[Connectivity Enhanced Safe Neural Network Planner for Lane Changing in Mixed Traffic]{Connectivity Enhanced Safe Neural Network Planner \\for Lane Changing in Mixed Traffic}

\author{Xiangguo Liu}
\orcid{0000-0002-1944-3923}
\affiliation{%
  \institution{Northwestern University}
  \city{Evanston}
  \state{IL}
  \country{USA}
  \postcode{60201}
}
\email{xg.liu@u.northwestern.edu}

\author{Ruochen Jiao}
\affiliation{%
  \institution{Northwestern University}
  \city{Evanston}
  \state{IL}
  \country{USA}}
\email{RuochenJiao2024@u.northwestern.edu}

\author{Bowen Zheng}
\affiliation{%
 \institution{Pony.ai}
 \city{Fremont}
 \state{CA}
 \country{USA}}
\email{bowen.zheng@pony.ai}

\author{Dave Liang}
\affiliation{%
 \institution{Pony.ai}
 \city{Fremont}
 \state{CA}
 \country{USA}}
\email{dave.liang@pony.ai}

\author{Qi Zhu}
\affiliation{%
  \institution{Northwestern University}
  \city{Evanston}
  \state{IL}
  \country{USA}
}
\email{qzhu@northwestern.edu}

\begin{abstract}
Connectivity technology has shown great potentials in improving the safety and efficiency of transportation systems by providing information beyond the perception and prediction capabilities of individual vehicles. However, it is expected that human-driven and autonomous vehicles, and connected and non-connected vehicles need to share the transportation network during the transition period to fully connected and automated transportation systems. Such mixed traffic scenarios significantly increase the complexity in analyzing system behavior and quantifying uncertainty for highly interactive scenarios, e.g., lane changing. It is even harder to ensure system safety when neural network based planners are leveraged to further improve efficiency. In this work, we propose a connectivity-enhanced neural network based lane changing planner. By cooperating with surrounding connected vehicles in dynamic environment, our proposed planner will adapt its planned trajectory according to the analysis of a safe evasion trajectory. We demonstrate the strength of our planner design in improving efficiency and ensuring safety in various mixed traffic scenarios with extensive simulations. We also analyze the system robustness when the communication or coordination is not perfect.
\end{abstract}


\newcommand{\BibTeX}{\rm B\kern-.05em{\sc i\kern-.025em b}\kern-.08em\TeX}

\begin{document}


\pagestyle{fancy}
\fancyhead{}


\maketitle 


\section{Introduction}




Connectivity technology for autonomous driving has been increasingly studied in academia and industry, which is expected to significantly improve safety~\cite{ali2019hazard}, energy efficiency and time efficiency~\cite{liu2021trajectory,liu2022markov} of transportation systems. In a connectivity-enhanced transportation system, vehicles can communicate with each other and/or surrounding infrastructures via dedicated short range communication (DSRC)~\cite{kenney2011DSRC} or cellular vehicle-to-everything (C-V2X)~\cite{garcia2021tutorial}. These communications share important information of vehicles' current states (e.g., location, speed, acceleration) and future intentions (e.g., planned actions and trajectories) that go well beyond the perception and prediction capabilities of individual vehicles, e.g., sharing information that are out of sight of the ego vehicle or intentions that cannot be accurately predicted~\cite{jiao2022semi}. Beyond that, vehicles can negotiate and coordinate in distributed manner~\cite{wang2019cooperative}, or follow instructions from a central unit~\cite{zheng2019design} to further optimize the system. This makes connectivity a great complementary to both autonomous vehicles and human-driven vehicles. 

However, the adoption of connectivity technology faces significant challenges from the increasing complexity in analyzing system behavior and ensuring its safety, especially with the wider usage of neural network based components in vehicle decision making~\cite{zhu2021safety}. During the transition period to the next-generation transportation system, a mixed traffic stream of human-driven and autonomous vehicles~\cite{zhang2022trust}, and connected and non-connected vehicles need to share the transportation network. Recent progress has been made to ensure system safety when all vehicles are of the same type~\cite{khayatian2018rim,fu2018infrastructure}, or connectivity is not enabled~\cite{pek2020using,shalev2017formal,li2020adaptive,liu2022neural}, and only some works~\cite{pek2020using,liu2022neural} can generalize to systems with neural network based components. Lane changing, especially in congested traffic, is a challenging scenario, in which those vehicles need to closely interact with each other~\cite{liu2020impact}. It is an open challenge to model the behavior of a general system with all kinds of vehicles, not to mention providing safety guarantee while not overly sacrificing efficiency.

\begin{figure*}[tbp]
\centering\includegraphics[scale=0.52]{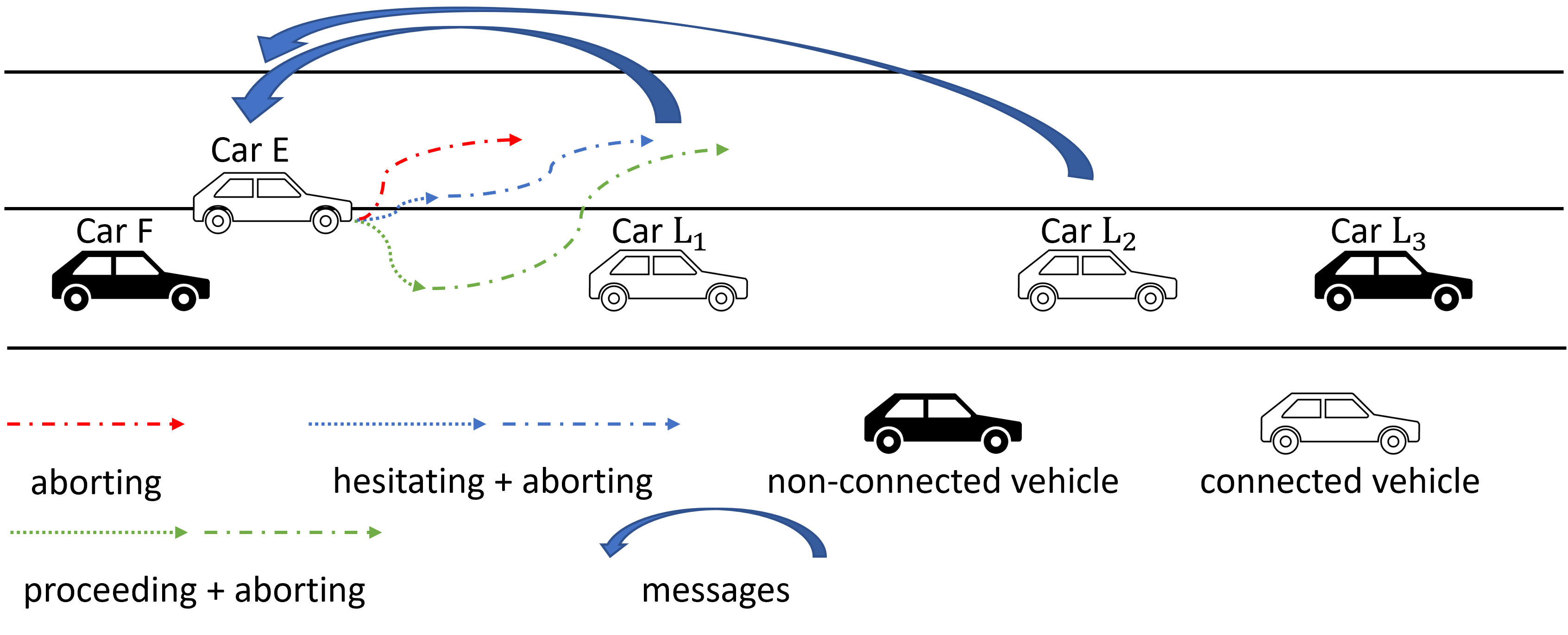}
\caption{The ego vehicle $E$ intends to change lane. With connectivity technology, vehicle $E$ receives planned acceleration profiles and real-time motion states from connected leading vehicles $L_i$, $1 \leq i \leq N$, and then analyzes the maximum deceleration of vehicle $L_1$ during the lane changing process. By identifying the behavior of following vehicle $F$ and analyzing system safety in the worst case, vehicle $E$ may proceed to change lane, hesitate around the current lateral position, or abort the lane changing plan. This figure shows an example with $N=2$ and the following vehicle $F$ is non-connected.}
\label{fig:cv_scenario}
\end{figure*}

To overcome these challenges, we propose a connectivity-enhanced neural network based planner design, which can ensure safety for lane changing in mixed traffic. In this general framework, we assume that the ego vehicle is connected and autonomous, while surrounding vehicles can be either connected or non-connected, and either autonomous or human-driven\footnote{For connected human-driven vehicles, we assume that a larger inter-vehicle distance needs to be kept to ensure safety, as human-driven vehicles can have larger execution error compared with autonomous vehicles.}. Figure~\ref{fig:cv_scenario} shows an example of the lane changing scenario. The connected ego vehicle $E$ intends to change to the target lane. From the neighbor area to the traffic downstream, there are a set of connected leading vehicles $L_1$, $L_2$, $\cdots$, $L_N$ and one non-connected leading vehicle $L_{N+1}$ in the target lane. In this work, we are considering all possible scenarios with varied type and number of surrounding vehicles in our planner design. Depending on the dynamic mixed traffic environment, the number of connected leading vehicles $N$ may vary\footnote{By $N=0$, it represents that the immediate leading vehicle is non-connected.}. The following vehicle $F$ in the target lane can be either connected or non-connected. 

To ensure safety, we verify the existence of a safe evasion trajectory following the planned trajectory in the worst case. That is, the planned trajectory does not need to complete the lane changing in one go. As long as there is a safe evasion trajectory, the ego vehicle can start the attempt and interact with surrounding vehicles. We analyze the worst case for the ego vehicle in dynamical environments. For the case without a connected leading vehicle, i.e., $N=0$, we can directly assume that in the worst case, vehicle $L_1$ takes the maximum deceleration under the mechanical constraints. In other cases that $N>0$, we can leverage the coordination from connected vehicles to prevent overly conservative planning. We evaluate the fastest evasion trajectory of the ego vehicle in emergency situations. For instance, when there is an emergency brake in the downstream, connected vehicles that first realize the event can collaboratively take a smaller deceleration, if it is safe for themselves, thus leaving more reaction time for the ego vehicle. If the evasion trajectory can prevent collisions, the ego vehicle can proceed to change lanes under the neural network based planners, otherwise it has to hesitate around the boundary of the two lanes or return back to the original lane for safety.

\smallskip
\noindent
The contribution of our work can be summarized as:
\begin{myitemize}
\item We propose a connectivity-enhanced neural network based lane changing planner in mixed traffic environment with human-driven and autonomous vehicles, and connected and non-connected vehicles.
\item Surrounding vehicles' behaviors are modeled via connectivity or aggressiveness assessment for safety analysis. Our planner design is guaranteed to be safe if the communication is perfect, the aggressive assessment is accurate, or if we choose to always treat the following vehicle as aggressive.
\item We demonstrate significant system efficiency improvements by leveraging connectivity in dynamic environment, through simulations with comprehensive experimental settings. We also analyze system robustness when coordination between connected vehicles is not perfect.
\end{myitemize}

The rest of this paper is organized as follows. In Section~\ref{sec: background and related work}, we review related works on lane changing, inter-vehicle interaction, neural network-based planning and connected vehicles. In Section~\ref{sec: methodology}, we present our connectivity-enhanced planning framework. Section~\ref{sec: experiments} shows the experimental results and Section~\ref{sec: conclusion} concludes the paper.

\section{Related Work}\label{sec: background and related work}


Planner design in autonomous driving has been a very active research area for its potential impact in greatly improving transportation system performance and also its challenge in ensuring system safety at the same time. There is an urgent need to prevent overly conservative design while also provide safety guarantees when considering environment uncertainty and complex inter-vehicle interactions~\cite{zhan2016non,liu2022neural,fisac2018probabilistically}. The work in~\cite{pek2020using} proposes a concept of legal safety, which means that autonomous vehicles will not be the cause of accidents and there is no collision if surrounding vehicles obey traffic rules. This is realized by proving the existence of the fail-safe trajectory under the planner all the time. Similarly, the concept of responsibility-sensitive safety is proposed in~\cite{shalev2017formal}, which assumes that other participants behave according to common-sense rules and defines appropriate responses of autonomous vehicles in near-accident scenarios. However, safety can be compromised if other vehicles' behaviors violate the assumptions. The works in~\cite{zhan2016non,li2020adaptive} develop a non-conservatively defensive driving strategy, which leverages sampling based or optimization based methods. Planned trajectory is executed after the safety evaluation. Model Predictive Control (MPC) based approaches can address safety issues by applying constraints or designing cost functions~\cite{Wu2020Amphibious,xiao2022robotic}.

Machine learning based techniques are increasingly popular in planning and decision making for autonomous driving~\cite{mirchevska2018high,jiao2022tae,zhang2022attention}, for their potential in improving average system performance under complex scenarios. The work in~\cite{sun2019behavior} proposes a concept of social perception, which inferences surrounding environment from other vehicles' reactions. It then leverages inverse reinforcement learning (IRL) to acquire cost function of human driving, and uses Markov Decision Process (MDP) to get probabilistically optimal solutions. \cite{li2020interaction} formulates the lane changing planning problem as a partially observable Markov Decision Process (POMDP), in which the cooperativeness of other traffic participants is an unobservable state. It predicts future actions of human cars via logistic regression classifier, and solves the POMDP by Monta-Carlo Tree Search. The work in~\cite{cao2020reinforcement} proposes a hierarchical reinforcement and imitation learning (H-REIL) approach specifically for near-accident scenarios, which consists of low-level policies learned by imitation learning (IL) for discrete driving modes, and a high-level policy learned by reinforcement learning (RL) that switches between different driving modes. \cite{liu2022physics} proves the strength of a hierarchical neural network based planner regarding safety and verifiability, compared with a single neural network based planner.

Although these methods demonstrate great performance improvement, it is still quite challenging to verify the safety for learning-enabled systems~\cite{Wang2020Energy,wang2022design}. System efficiency is also restricted by uncertainties from perception~\cite{orzechowski2018tackling} and prediction results of individual vehicles. And connectivity can enhance the transportation system by reducing such  uncertainties~\cite{xing2019personalized,park2018enhancing,aoki2020cooperative}. For instance, the work in~\cite{khayatian2018rim} proposes an intersection management scheme, in which the central manager assigns arriving speed and arriving time to vehicles. 
It assumes that all vehicles are connected and autonomous. The approach in~\cite{fu2018infrastructure} leverages Dynamic Bayesian Networks (DBNs) to model vehicle state evolution. The central manager can send out warning messages to vehicles for collision avoidance. It assumes that all vehicles are connected and human-driven, and collision rate depends on velocity and driver reaction time. The work in~\cite{dong2020facilitating} assumes that all vehicles are connected, and leverages deep reinforcement learning (DRL) for behavior-level decision making in lane changing. In particular, it gets performance improvement by incorporating traffic status in the downstream with vehicle-to-vehicle communication.

There are several works developed for mixed traffic~\cite{zhou2022multi,liu2021proactive,du2021cooperative,Chen2023Mixed} of connected and non-connected vehicles. The work in~\cite{ha2020leveraging} presents an RL-based multi-agent longitudinal planner for connected and autonomous vehicles, which adjusts speeds in upstream traffic to mitigate traffic shock-waves downstream. The results suggest that even for a penetration rate of 10\%, connected and autonomous vehicles can significantly mitigate bottlenecks in highway traffic. The approach in~\cite{nassef2020building} leverages RL for trajectory recommendation to the connected vehicles in highway merging scenarios. It assumes that not all vehicles are connected and uses camera in roadside for data fusion, in order to map all vehicles. The work in~\cite{han2020behavior} proposes an RL-based method for connected and autonomous vehicles to decide actions such as whether to change lane or keep lane based on the observation and shared information from neighbors. The system is modeled by hybrid partially observable Markov Decision Process (HPOMDP) as not all vehicles are connected. However, it does not explicitly model inter-vehicle interaction, and the safety highly depends on accurate modeling of the surrounding vehicles, especially non-connected vehicles. In this work, we propose a general lane changing planner with safety guarantee in mixed traffic, which can work safely and efficiently under any penetration rate of connected vehicles.

\section{Design of Our Framework}
\label{sec: methodology}

\subsection{Overview}
\begin{figure*}[!t]
\centering\includegraphics[scale=0.45]{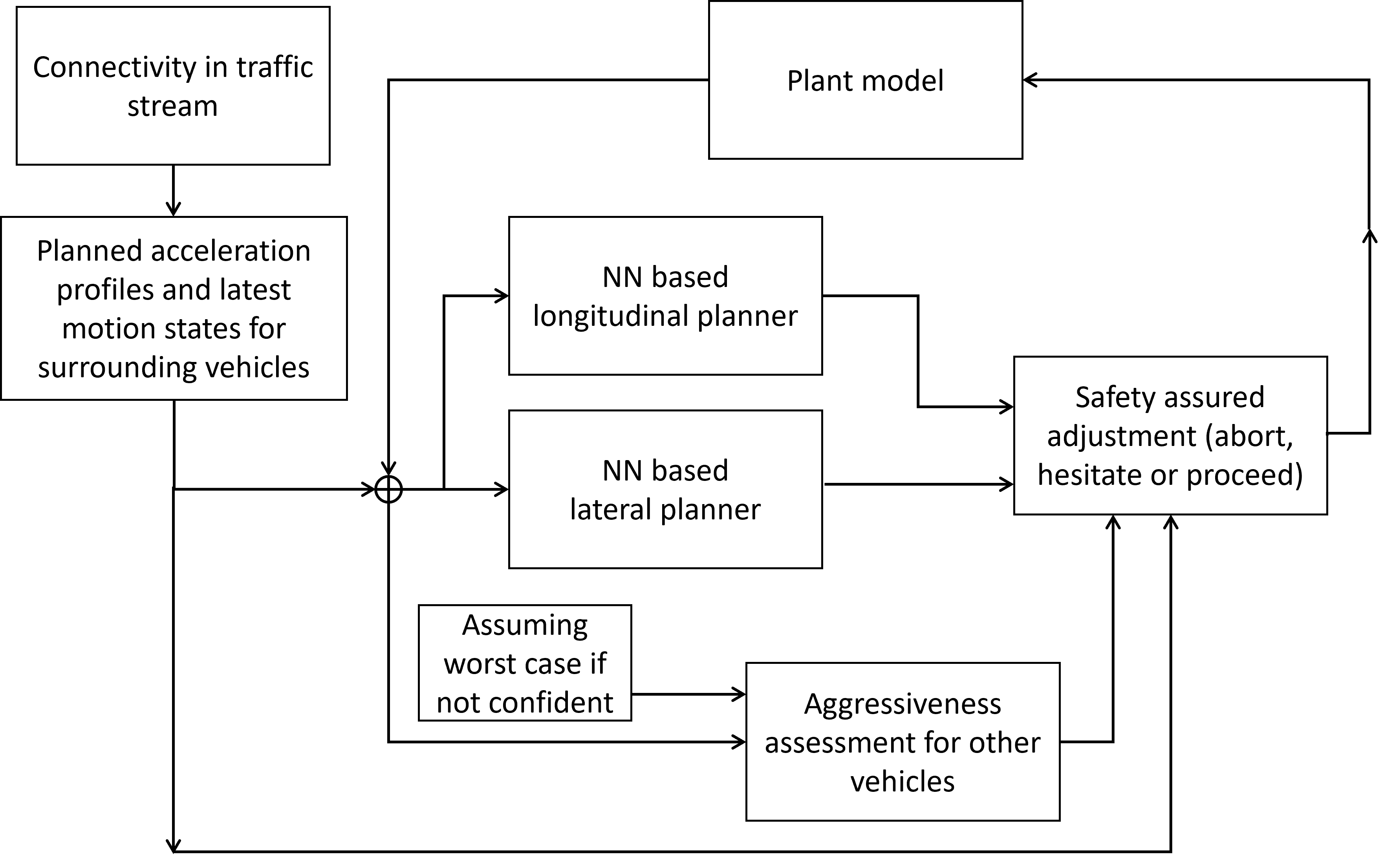}
\caption{By incorporating the planned acceleration profiles and the real-time motion states of surrounding vehicles updated by connectivity technology, we can further improve the performance of neural network based lane changing planner. At the same time, we can derive the maximum deceleration of the leading vehicle $L_1$ (scenario as shown in Figure~\ref{fig:cv_scenario}), perform aggressiveness assessment and system analysis for the worst case, and adjust trajectory to ensure safety.}
\label{fig:cv_framework}
\end{figure*}

By leveraging the connectivity technology, our planner design can further improve system efficiency while ensuring system safety at the same time. The framework is presented in Figure~\ref{fig:cv_framework}. Based on the planned acceleration profiles in the planning horizon and the real-time motion states of surrounding vehicles, we can leverage neural networks for longitudinal and lateral trajectory planning. At the same time, we can derive the maximum deceleration of the leading vehicle $L_1$ (scenario as shown in Figure~\ref{fig:cv_scenario}), and then perform system analysis for the worst case and adjust trajectory to ensure safety. Here we adopt the same aggressiveness assessment method for the following vehicle $F$ as in~\cite{liu2022neural} when vehicle $F$ is non-connected. For the case that the following vehicle $F$ is connected, we assume that it is collaborative.

Safety analysis and trajectory adjustment are conducted periodically. At every step during the lane changing, the ego vehicle has three behavior-level options with strictly decreasing preference: proceed changing lane, hesitate around current lateral position, or abort changing lane and return back to the original lane. It analyzes the state after executing the accelerations under neural network based planners for one time step. If it has a safe evasion trajectory in the worst case, it can go ahead and change lanes. Otherwise, it has to attempt a less preferred behavior until a safe evasion trajectory is found following that.


\subsection{Connectivity Assumptions}
In this work, we assume that the connected leading vehicles $L_i$, $1 \leq i \leq N$ (for example, $N=2$ in the scenario from Figure~\ref{fig:cv_scenario}) will collaboratively assist the lane changing process of the ego vehicle $E$ and prevent collision with their own immediate leading vehicle $L_{i+1}$. Specifically, the connected leading vehicle $L_i$ will keep its acceleration within the range $[-a_i^{m,d}, a_i^{m,a}]$ to execute its own driving task if that is sufficient for keeping a safe distance between $L_i$ and $L_{i+1}$. It is noted that the values of $a_i^{m,d}$ and $a_i^{m,a}$ will be communicated to other connected vehicles. Only in emerging scenarios, e.g., when the leading vehicle $L_{i+1}$ decelerates suddenly, $L_i$ can violate such \emph{`promise'} and take a deceleration $a_{x,i,d}>a_i^{m,d}$ to ensure safety. Note that this is considered as an \emph{expected violation} of the promise, and our planner framework can ensure safety in such cases (later in Section~\ref{sec: experiments}, we will demonstrate the impact when such promise is violated \emph{unexpectedly}, e.g., when one surrounding vehicle is malfunctioning or influenced by other unknown obstacles). 

To generalize the model, we assume that both connected human-driven vehicles and connected autonomous vehicles communicate their planned acceleration range to other vehicles in the same manner. However, human-driven vehicles could have larger acceleration range because there is typically larger uncertainty in human driver's behavior and execution process. For non-connected vehicles, we assume that it can take any acceleration value within their mechanical constraints.

\subsection{Safety Analysis}

Next we will analyze the evasion trajectory given the system states. Assuming that vehicle $L_i$ decelerates with $a_{x,i,d}$, vehicle $L_{i-1}$ needs to decelerate with $z_{x,i-1,d}$ to prevent collisions. By letting $p_m$ denote the minimum safe distance between two vehicles, we have

\begin{equation}\label{eq:leading vehicle decelerates_0}
\begin{cases}
p_{x,i}+v_{x,i} t_{L_i}+\frac{a_{x,i,d} t_{L_i}^2}{2} - p_{x,i-1} - v_{x,i-1} t_{L_i} - \frac{z_{x,i-1,d} t_{L_i}^2}{2} - p_{m}=0
\\
\qquad \qquad \text{if } v_{x,i} < v_{x,i-1} \text{ and } \frac{v_{x,i}}{a_{x,i,d}} \geq  \frac{v_{x,i-1}}{z_{x,i-1,d}},

\\
p_{x,i}+\frac{v_{x,i}^2}{2 a_{x,i,d}} - p_{x,i-1}-\frac{v_{x,i-1}^2}{2 z_{x,i-1,d}} - p_{m}=0
\\
\qquad \qquad \text{otherwise},
\end{cases}
\end{equation}
where $p_{x,i}$, $v_{x,i}$, $p_{x,i-1}$ and $v_{x,i-1}$ denote the position and velocity of vehicle $L_i$ and $L_{i-1}$, respectively. The first equation corresponds to the case that two vehicles can have the same velocity at $t_{L_i}=\frac{v_{x,i-1}-v_{x,i}}{z_{x,i-1,d}-a_{x,i,d}}$ and reach minimum distance gap at the same time. The second equation corresponds to the case that $L_{i-1}$ is the closest to $L_i$ when it stops.

Then we can obtain the deceleration of vehicle $L_{i-1}$ by solving Equation~\eqref{eq:leading vehicle decelerates_0}
\begin{equation}\label{eq:leading vehicle decelerates_1}
z_{x,i-1,d}=
\begin{cases}
\frac{3 (v_{x,i-1}-v_{x,i})^2}{2 (p_{x,i}-p_{x,i-1}-p_{m})}+a_{x,i,d}
\\
\qquad \qquad \text{if } v_{x,i} < v_{x,i-1} \text{ and } \frac{v_{x,i}}{a_{x,i,d}} \geq  \frac{v_{x,i-1}}{z_{x,i-1,d}},
\\
\frac{v_{x,i-1}^2}{2 (p_{x,i}-p_{x,i-1}-p_{m}) + \frac{v_{x,i}^2}{a_{x,i,d}}}
\\
\qquad \qquad \text{otherwise}.
\end{cases}
\end{equation}

Considering that vehicle $L_{i-1}$ has claimed in a message that it will take an acceleration in range $[-a_{i-1}^{m,d}, a_{i-1}^{m,a}]$ to execute its own driving task, we then have the deceleration of vehicle $L_{i-1}$ in the worst case

\begin{equation}\label{eq:leading vehicle decelerates_2}
a_{x,i-1,d}=max(a_{i-1}^{m,d}, z_{x,i-1,d}).
\end{equation}

Let $a_{x,d}$ and $a_{x,a}$ denote the maximal longitudinal deceleration and acceleration of a vehicle under mechanical constraints. Let $a_{y,m}$ denote the absolute value of the maximal lateral acceleration under mechanical constraints. In the worst case, we assume that the non-connected leading vehicle $L_{N+1}$ can decelerate with $a_{x,N+1,d}=a_{x,d}$, and then we can obtain $a_{x,N,d}$ with Equations~\eqref{eq:leading vehicle decelerates_1} and~\eqref{eq:leading vehicle decelerates_2}. Similarly, we can compute the decelerations $a_{x,N-1,d}$, $a_{x,N-2,d}$, $\cdots$, $a_{x,1,d}$ for the other leading vehicles in the worst case.

Given that the leading vehicle $L_1$ decelerates with $a_{x,1,d}$, its position at time $t$ is
\begin{equation}\label{eq:leading vehicle decelerates_3}
p_{x,1,t}=
\begin{cases}
p_{x,1}+v_{x,1} t - \frac{a_{x,1,d} t^2}{2}
\\
\qquad \qquad \text{if } t \leq \frac{v_{x,1}}{a_{x,1,d}},
\\
p_{x,1}+\frac{v_{x,1}^2}{2 a_{x,1,d}}
\\
\qquad \qquad \text{otherwise}.
\end{cases}
\end{equation}

Next we will analyze the evasion trajectory for the ego vehicle $E$ in the worst case. If it is safe, the ego vehicle $E$ can proceed to change lane under the neural network based planners, otherwise it can adapt behavior and trajectory to ensure safety. For the lateral motion, the fastest evasion trajectory is to accelerate with $a_y=-a_{y,m}$ when $t \in [0, t_1]$ and then $a_y=a_{y,m}$ when $t \in [t_1, t_{y,f}]$. It 
can be obtained by solving the equation below.
\begin{equation}\label{eq:lateral time_0}
\begin{cases}
p_{y,t_0}+v_{y,t_0} t_1 - \frac{a_{y,m} t_1^2}{2}+(v_{y,t_0}-a_{y,m} t_1) (t_{y,f}-t_1) \\ \qquad + \frac{a_{y,m}(t_{y,f}-t_1)^2}{2}=\frac{w_l-w_v}{2},
\\
v_{y,t_0}-a_{y,m} t_1 + a_{y,m}(t_{y,f}-t_1)=0,
\end{cases}
\end{equation}
where $p_{y,t_0}$ and $v_{y,t_0}$ are the lateral position and velocity of the ego vehicle when $t=0$. 
The centers of the original and target lane are $y=0$ and $y=w_l$, respectively. The width of a vehicle is $w_v$.

As for the longitudinal motion, the optimal trajectory is analyzed intuitively when the immediate leading vehicle $L_1$ decelerates with $a_{x,d}$~\cite{liu2022neural}. It is that the ego vehicle first accelerates with $a_{x,a}$ and then decelerates with $a_{x,d}$, and keeps the distance gap no smaller than $p_m$ when $t \in [0, t_{y,f}]$. It is the fastest longitudinal motion to get closer to the leading vehicle, thus obtaining more time for the lateral evasion before the following vehicle catches up. 

However, in this work, it is a more general situation that the leading vehicle $L_1$ can have deceleration $a_{x,1,d} \leq a_{x,d}$. Thus when the ego vehicle $E$ decelerates to the same velocity with the leading vehicle $L_1$, its deceleration should change from $a_{x,d}$ to $a_{x,1,d}$. Otherwise the headway of the ego vehicle will increase when its velocity is smaller than that of the vehicle $L_1$. In that way, the ego vehicle is overreacting for collision avoidance, which cannot lead to an optimal trajectory.

\begin{figure*}[!t]
\centering\includegraphics[scale=0.5]{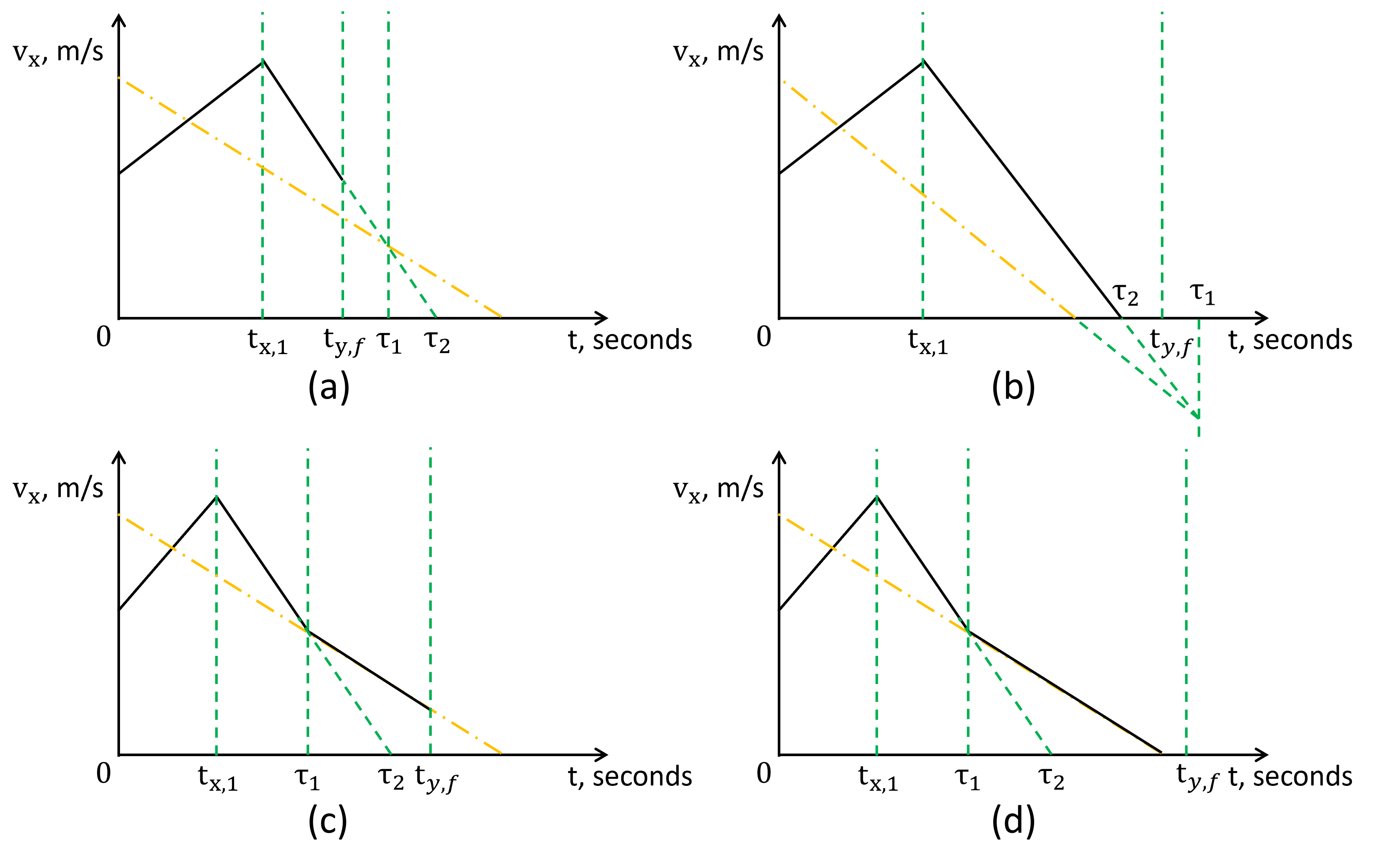}
\caption{Depending on the initial states and $t_{y,f}$, there are four cases of longitudinal evasion trajectory for the ego vehicle $E$, which are presented in four sub-figures. The black solid line and yellow dot dash line represent velocity curves of the ego vehicle $E$ and the leading vehicle $L_1$, respectively. (a) represents that the ego vehicle has already finished its lateral motion before it decelerates to the same velocity with vehicle $L_1$ and changes its deceleration. (b) is that the ego vehicle decelerates with $a_{x,d}$ until it stops, and then keeps $v_x=0$ until $t=t_{y,f}$. (c) is that the ego vehicle has already finished its lateral motion before it reaches $v_x=0$ with deceleration $a_{x,1,d}$. (d) represents that the ego vehicle decelerates with $a_{x,1,d}$ until it stops, and then keeps $v_x=0$ until $t=t_{y,f}$.
}
\label{fig:cvs1234}
\end{figure*}


There are three phases in the optimal longitudinal motion. The ego vehicle first accelerates with $a_x=a_{x,a}$ when $t \in [0, t_{x,1}]$, and then decelerates with $a_x=-a_{x,d}$ when $t \in [t_{x,1}, t_{x,2}]$, and then decelerates with $a_x=-a_{x,1,d}$ until it stops when $t \in [t_{x,2}, t_{y,f}]$. The position of the ego vehicle when $t=t_{y,f}$ is formulated as 

\begin{equation}\label{eq:leading vehicle decelerates_4}
p_{x,t_{y,f}}=
\begin{cases}
p_{x,t_0}+v_{x,t_0}t_{x,1}+\frac{a_{x,a}t_{x,1}^2}{2}+(v_{x,t_0}+a_{x,a}t_{x,1})(t_{y,f}-t_{x,1}) \\ \qquad -\frac{a_{x,d}(t_{y,f}-t_{x,1})^2}{2}
\\
\qquad \qquad \text{if } v_{x,t_0}+a_{x,a}t_{x,1}-a_{x,d}(t_{y,f}-t_{x,1}) \\ \qquad \qquad \qquad \geq max(0, v_{x,1}-a_{x,1,d} t_{y,f}),
\\
p_{x,t_0}+v_{x,t_0}t_{x,1}+\frac{a_{x,a}t_{x,1}^2}{2}+ \frac{(v_{x,t_0}+a_{x,a}t_{x,1})^2}{2 a_{x,d}}
\\
\qquad \qquad \text{else if } \frac{v_{x,1}}{a_{x,1,d}} \leq \frac{v_{x,t_0}+a_{x,a}t_{x,1}}{a_{x,d}}+t_{x,1} \leq t_,{y,f},
\\
p_{x,t_0}+v_{x,t_0}t_{x,1}+\frac{a_{x,a}t_{x,1}^2}{2}+(v_{x,t_0}+a_{x,a}t_{x,1})(t_{x,2}-t_{x,1}) \\ \qquad -\frac{a_{x,d}(t_{x,2}-t_{x,1})^2}{2} -\frac{a_{x,1,d}(t_{y,f}-t_{x,2})^2}{2}
\\ \qquad
+(v_{x,t_0}+a_{x,a}t_{x,1}-a_{x,d}(t_{x,2}-t_{x,1}))(t_{y,f}-t_{x,2}) 
\\
\qquad \qquad \text{else if } \frac{v_{x,1}}{a_{x,1,d}} \geq t_{y,f},
\\
p_{x,t_0}+v_{x,t_0}t_{x,1}+\frac{a_{x,a}t_{x,1}^2}{2}+(v_{x,t_0}+a_{x,a}t_{x,1})(t_{x,2}-t_{x,1})
\\ \qquad
-\frac{a_{x,d}(t_{x,2}-t_{x,1})^2}{2}+\frac{(v_{x,t_0}+a_{x,a}t_{x,1}-a_{x,d}(t_{x,2}-t_{x,1}))^2}{2 a_{x,1,d}} 
\\
\qquad \qquad \text{otherwise},
\end{cases}
\end{equation}
where $p_{x,t_0}$ and $v_{x,t_0}$ are the longitudinal position and the velocity of the ego vehicle when $t=0$, respectively. 

Depending on the initial states and $t_{y,f}$, there are four cases in Equation~\eqref{eq:leading vehicle decelerates_4}, which correspond to the four velocity curves in Figure~\ref{fig:cvs1234}. The black solid line and yellow dot dash line represent velocity curves of the ego vehicle $E$ and the leading vehicle $L_1$, respectively. The first case represents that the ego vehicle has already finished its lateral motion before it decelerates to the same velocity with vehicle $L_1$ and changes its deceleration. The second case is that the ego vehicle decelerates with $a_{x,d}$ until it stops, and then keeps $v_x=0$ until $t=t_{y,f}$. The third case is that the ego vehicle has already finished its lateral motion before it reaches $v_x=0$ with deceleration $a_{x,1,d}$. The fourth case represents that the ego vehicle decelerates with $a_{x,1,d}$ until it stops, and then keeps $v_x=0$ until $t=t_{y,f}$.


Given that the two vehicles reach the same velocity when $t=t_{x,2}$, we have $v_{x,t_0}+a_{x,a}t_{x,1}-a_{x,d}(t_{x,2}-t_{x,1})=v_{x,1}-a_{x,1,d} t_{x,2}$. In all four cases, the distance gap between the ego vehicle $E$ and the leading vehicle $L_1$ reaches the minimum when $t=t_{y,f}$. By letting $p_{x,1,t_{y,f}}-p_{x,t_{y,f}}-p_m=0$, we can compute the value of $t_{x,1}$ as 

\begin{equation}\label{eq:leading vehicle decelerates_5}
t_{x,1}=
\begin{cases}
t_{y,f}-\sqrt{t_{y,f}^2+\frac{2v_{x,t_0}t_{y,f} - a_{x,d}t_{y,f}^2 + 2p_{x,t_0} + 2p_m-2p_{x,1,t_{y,f}} }{a_{x,a}+a_{x,d}}}
\\
\qquad \qquad \text{if } v_{x,t_0}+a_{x,a}t_{x,1}-a_{x,d}(t_{y,f}-t_{x,1}) 
\\ \qquad \qquad \qquad \geq max(0, v_{x,1}-a_{x,1,d} t_{y,f}),
\\
-\frac{v_{x,t_0}}{a_{x,a}}+\sqrt{(\frac{v_{x,t_0}}{a_{x,a}})^2-\frac{2p_{x,t_0}a_{x,d}+v_{x,t_0}^2+2 p_m a_{x,d}-2 p_{x,1,t_{y,f}} a_{x,d}}{(a_{x,a}+a_{x,d})a_{x,a}}}
\\
\qquad \qquad \text{else if } \frac{v_{x,1}}{a_{x,1,d}} \leq \frac{v_{x,t_0}+a_{x,a}t_{x,1}}{a_{x,d}}+t_{x,1} \leq t_{y,f},
\\
\frac{-(v_{x,t_0}-v_{x,1}) + \sqrt{(v_{x,t_0}-v_{x,1})^2-2(a_{x,a}+a_{x,1,d})C_2}}{a_{x,a}+a_{x,1,d}}
\\
\qquad \qquad \text{else if } \frac{v_{x,1}}{a_{x,1,d}} \geq t_{y,f},
\\
\frac{-(v_{x,t_0}-v_{x,1}) + \sqrt{(v_{x,t_0}-v_{x,1})^2-2(a_{x,a}+a_{x,1,d})C_2}}{a_{x,a}+a_{x,1,d}}
\\
\qquad \qquad \text{otherwise},
\end{cases}
\end{equation}
where $C_2=\frac{(v_{x,t_0}-v_{x,1})^2 + (2p_{x,t_0}+2p_m-2p_{x,1})(a_{x,d}-a_{x,1,d})}{2 (a_{x,a}+a_{x,d})}$.




It is noted that if $t_{y,f}^2+\frac{2v_{x,t_0}t_{y,f} - a_{x,d}t_{y,f}^2 + 2p_{x,t_0} + 2p_m-2p_{x,1,t_{y,f}} }{a_{x,a}+a_{x,d}}\leq 0$, the ego vehicle can keep accelerating until $t=t_{y,f}$, i.e., $t_{x,1}=t_{y,f}$, and remain safe. If $2v_{x,t_0}t_{y,f} - a_{x,d}t_{y,f}^2 + 2p_{x,t_0} + 2p_m-2p_{x,1,t_{y,f}}>0$ in the first case, or $2p_{x,t_0}a_{x,d}+v_{x,t_0}^2+2 p_m a_{x,d}-2 p_{x,1,t_{y,f}} a_{x,d}>0$ in the second case, or $C_2>0$ in the third and fourth cases, we have $t_{x,1}<0$, which means the ego vehicle cannot prevent collisions even when it keeps decelerating with $a_{x,d}$ from $t=0$, which means that the safe evasion trajectory does not exist.


Let $\tau_1$ denote the time that the ego vehicle decelerates to the same velocity with the leading vehicle, and let $\tau_2$ denote the time that the ego vehicle longitudinally decelerates to a velocity of zero. We present $\tau_1$ and $\tau_2$ in all four cases in Figure~\ref{fig:cvs1234}. Based on the value of $t_{x,1}$, we can obtain the position of the ego vehicle at time $t \in [0, t_{y,f}]$ as

\begin{equation}\label{eq:leading vehicle decelerates_6}
p_{x,t}=
\begin{cases}
p_{x,t_0}+v_{x,t_0}t+\frac{a_{x,a}t^2}{2}
\\
\qquad \qquad \text{if } t \in [0, t_{x,1}],
\\
p_{x,t_0}+v_{x,t_0}t_{x,1}+\frac{a_{x,a}t_{x,1}^2}{2}+(v_{x,t_0}+a_{x,a}t_{x,1})(t-t_{x,1}) 
\\
\qquad
-\frac{a_{x,d}(t-t_{x,1})^2}{2}
\\
\qquad \qquad \text{else if } t \in [t_{x,1}, min(\tau_1, \tau_2, t_{y,f})],
\\
p_{x,1,t}-p_m
\\
\qquad \qquad \text{otherwise}.
\end{cases}
\end{equation}

Assuming that the following vehicle accelerates with $a_{x,f}\in [-a_{x,d}, a_{x,a}]$, its position at $t$ is
\begin{equation}\label{eq:leading vehicle decelerates_7}
p_{x,f,t}=
\begin{cases}
p_{x,f}+v_{x,f} t + \frac{a_{x,f} t^2}{2}
\\
\qquad \qquad \text{if } v_{x,f}+a_{x,f}t \geq 0,
\\
p_{x,f}+\frac{v_{x,f}^2}{2 |a_{x,f}|}
\\
\qquad \qquad \text{otherwise}.
\end{cases}
\end{equation}
In this work, we make the same assumptions as in~\cite{liu2022neural}: if the following vehicle is collaborative and willing to create gap for ego vehicle, it can at least decelerate with $a_{x,f}=-a_{x,d}$; if the following vehicle is aggressive, in the worst case, it can accelerate with $a_{x,f}=a_{x,a}$ to prevent the ego vehicle from cutting in. If $p_{x,t}-p_{x,f,t} \geq p_m$, $\forall t \in [0, t_{y,f}]$, the safe evasion trajectory exists, and is formulated in Equation~\eqref{eq:leading vehicle decelerates_6}. If it exists, the ego vehicle can go ahead and execute the planned trajectory, otherwise, it has to take a less preferred behavior, e.g., hesitate around the boundary of the two lanes or return back to the original lane for safety. It is noted that returning back to the original lane, which is the evasion trajectory computed in last control step, is already verified to be safe and feasible.

\section{Experimental Results}\label{sec: experiments}
\subsection{Effectiveness of Our Framework Under Perfect Coordination}
In this section, we first demonstrate the statistical performance of our connectivity-enhanced planner framework, and then present an example with detailed trajectories. We compare system performance under a few different planners to demonstrate the effectiveness of our approach: `CV\_all' represents the complete planner framework as in Figure~\ref{fig:cv_framework}, in which we leverage information shared by all connected vehicles; `CV\_follow' means that we only use information shared by the following connected vehicle $F$; `CV\_none' means that connectivity technology is not leveraged, which is the planner in~\cite{liu2022neural} and can be viewed as our main baseline; `No\_agg\_assess' means that we conservatively assume following vehicle is always aggressive and disable the aggressiveness assessment function.

In this work, we use synthesised data to train longitudinal and lateral planners, similarly as the work in~\cite{liu2022neural}. 
We conduct extensive simulations by uniformly sampling with $v_{x,t_0}\in[29, 31]$ meters per second and $\delta p \in [17, 22]$ meters, and setting $a_{i}^{m,d}=0.5$ meters per second squared, $v_{x,i}=30$ meters per second, $i=1, \cdots, N$. Here, $\delta p$ denotes the initial longitudinal inter-vehicle distance.

\begin{figure}[!t]
\centering\includegraphics[scale=0.5]{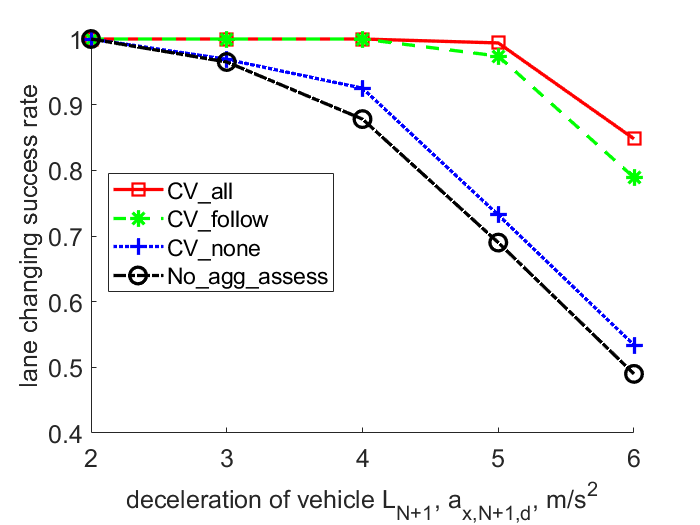}
\caption{Lane changing success rate under different planners are compared when the number of connected leading vehicles is $N=5$. The horizontal axes show the sudden deceleration of non-connected leading vehicle $L_{N+1}$.}
\label{fig:cvN5}
\end{figure}

Figure~\ref{fig:cvN5} presents the average lane changing success rate under different planners when the number of connected leading vehicles is $N=5$. The horizontal axes show the sudden deceleration of non-connected leading vehicle $L_{N+1}$. It is considered to be successful if the ego vehicle finally crosses the border of two lanes within the simulation horizon without any collision. Because safety is ensured by the trajectory adjustment function and there is indeed no collision in all simulations, we only present the results of lane changing success rate. 

As we expected, lane changing success rate decreases when the deceleration $a_{x, N+1, d}$ gets larger. Under these planners, `CV\_all', `CV\_follow' and `CV\_none' correspond to full, partial and none utilization of connectivity, respectively. `No\_agg\_assess' represents that when even aggressiveness assessment function is disabled, which leverages less information and is more conservative. It shows that `CV\_all' performs slightly better than `CV\_follow', and these two planners result in considerably larger success rate, when compared to `CV\_none' and `No\_agg\_assess'. \textbf{This clearly shows the effectiveness of our approach in improving system performance.} It means that understanding the following vehicle's intention can greatly help the lane changing maneuver of the ego vehicle, and connectivity of leading vehicles can further improve that.

\begin{table}[!t]
\centering
\caption{Lane changing success rate of different planners.}
\label{table: suc rate N_1_3_10}
\begin{tabular}{cc|cccc}
\hline
$N$                   & $a_{x,N+1,d}$ & CV\_all & CV\_follow & CV\_none & No\_agg\_assess \\ \hline
\multirow{5}{*}{1}  & 2              & 1       & 1          & 0.995    & 0.995           \\
                    & 3              & 1       & 1          & 0.875    & 0.83            \\
                    & 4              & 0.356   & 0.337      & 0.23     & 0.195           \\
                    & 5              & 0.001   & 0          & 0        & 0               \\
                    & 6              & 0       & 0          & 0        & 0               \\ \hline
\multirow{5}{*}{3}  & 2              & 1       & 1          & 1        & 1               \\
                    & 3              & 1       & 1          & 0.928    & 0.894           \\
                    & 4              & 0.998   & 0.982      & 0.712    & 0.672           \\
                    & 5              & 0.534   & 0.489      & 0.288    & 0.256           \\
                    & 6              & 0.01    & 0.008      & 0.024    & 0.022           \\ \hline
\multirow{5}{*}{10} & 2              & 1       & 1          & 1        & 1               \\
                    & 3              & 1       & 1          & 1        & 1               \\
                    & 4              & 1       & 1          & 0.993    & 0.998           \\
                    & 5              & 1       & 1          & 0.969    & 0.956           \\
                    & 6              & 1       & 1          & 0.95     & 0.928           \\ \hline
\end{tabular}
\end{table}

Table~\ref{table: suc rate N_1_3_10} shows the average lane changing success rate when $N$ and $a_{x, N+1, d}$ change. It consistently presents that the more utilization of connectivity, system performance is better. Moreover, when $N$ gets larger, it results in larger success rate because more connected leading vehicles can cooperate to leave larger space for the ego vehicle.

\begin{figure}[!t]
\centering\includegraphics[scale=0.5]{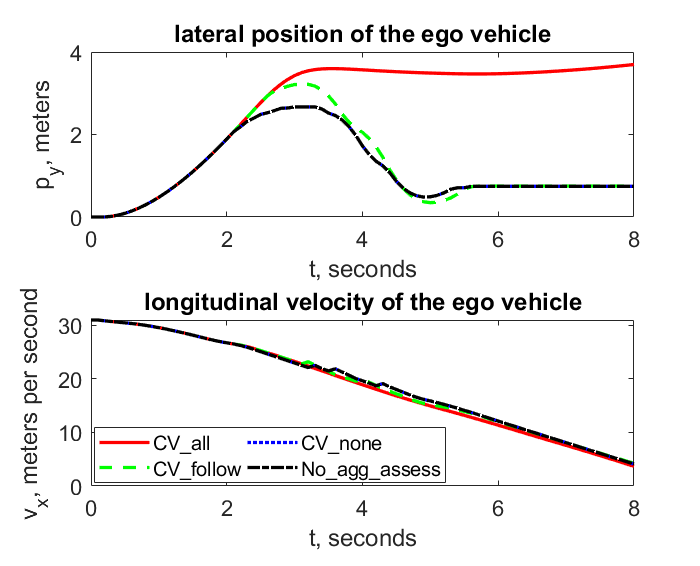}
\caption{Lateral position and longitudinal velocity of the ego vehicle in an example scenario are plotted under different planners, when the number of connected leading vehicles is $N=5$ and the deceleration of non-connected leading vehicle $L_{N+1}$ is $a_{x, N+1, d}=5$ meters per second squared.
}
\label{fig:cv_5_5_319}
\end{figure}

Figure~\ref{fig:cv_5_5_319} demonstrates the performance of our planner design with a specific example, in which $N=5$ and $a_{x, N+1, d}=5$ meters per second squared. It shows the lateral position and longitudinal velocity of the ego vehicle under different planners. Under `CV\_all' planner, the ego vehicle completes lane changing in about three seconds. Under other three planners, the ego vehicle first move laterally to the target lane, and then turn back to the original lane when $t=3$ seconds. `CV\_follow' results in longer time of staying in the target lane, compared with `CV\_none' and `No\_agg\_assess'. In this example, `CV\_none' and `No\_agg\_assess' lead to the same trajectory of the ego vehicle.

\begin{figure*}[!t]
\centering\includegraphics[scale=0.5]{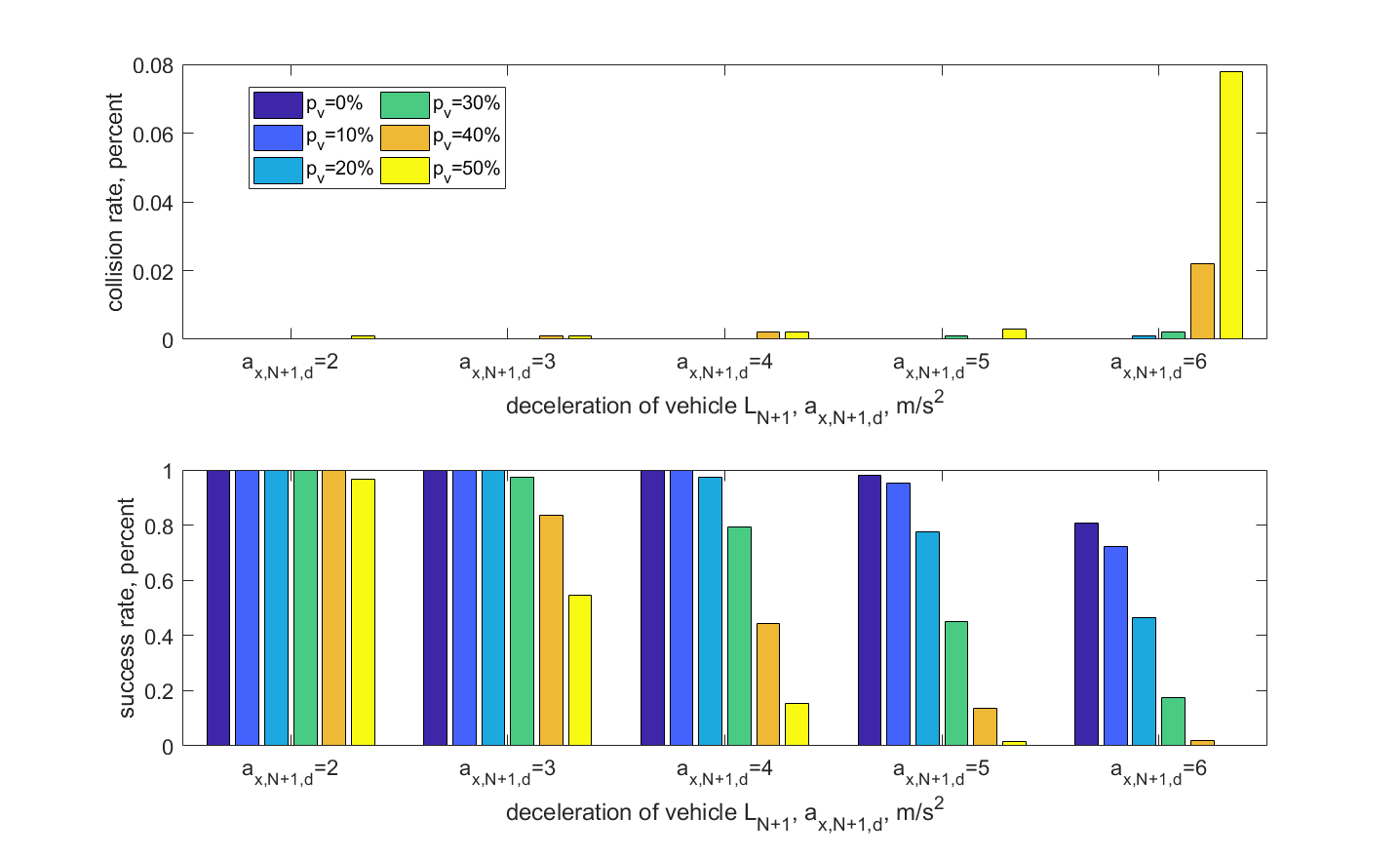}
\caption{Collision rate and lane changing success rate under different promise violation rates are presented when the number of connected leading vehicles is $N=10$. Promise violation rate $p_v$ is the probability that the promise is violated unexpectedly every control period. We assume that promise violation is independent among all connected vehicles, and the vehicle can take any deceleration following the uniform distribution $[z_{x,i,d}, a_{x,N+1,d}]$ when violating the promise. The horizontal axes show the sudden deceleration of non-connected leading vehicle $L_{N+1}$.
}
\label{fig:col_suc}
\end{figure*}

\subsection{Impact of Unexpected Promise Violation}

According to the \emph{promise} assumption introduced in Section~\ref{sec: methodology}, connected leading vehicle $L_i$ will keep its acceleration in the range $[-a_i^{m,d}, a_i^{m,a}]$ as long as that does not hurt its own safety. Let us define the promise violation rate $p_v$, which denotes the probability that the promise is violated unexpectedly in every control period. We assume that promise violation is independent among all connected vehicles, and the vehicle can take any deceleration following the uniform distribution $[z_{x,i,d}, a_{x,N+1,d}]$ when violating the promise.

Figure~\ref{fig:col_suc} presents the collision rate and lane changing success rate under varied promise violation rate $p_v$ when the first non-connected leading vehicle $L_{N+1}$ takes different sudden deceleration $a_{x,N+1,d}$. Generally speaking, a larger promise violation rate and deceleration result in smaller lane changing success rate and higher collision risk. However, when promise violation rate is within $20\%$ and the sudden deceleration is less than $5$ meters per second squared, collision rate is $0\%$ and success rate remains to be relatively high. \textbf{This shows the robustness of our approach.}

\section{Conclusion}\label{sec: conclusion}
In this work, we present a connectivity-enhanced planning framework for neural network based lane changing in mixed traffic. The framework can significantly improve lane changing performance by coordinating with surrounding connected vehicles in dynamic environment. Extensive experiments demonstrate the strength of our planner design in improving efficiency while ensuring safety. Our experiments suggest that (1) connectivity of the immediate following vehicle plays a more important role for ego vehicle's lane changing than the connectivity of leading vehicles, and (2) when there are more connected leading vehicles, the system performance can be further improved because more vehicles can coordinate to leave larger space for the ego vehicle. We also demonstrate the system robustness under different extent of promise violation rate of surrounding connected vehicles.

\pagebreak


\bibliographystyle{ACM-Reference-Format}
\bibliography{refs}


\begin{thebibliography}{42}


\ifx \showCODEN    \undefined \def \showCODEN     #1{\unskip}     \fi
\ifx \showDOI      \undefined \def \showDOI       #1{#1}\fi
\ifx \showISBNx    \undefined \def \showISBNx     #1{\unskip}     \fi
\ifx \showISBNxiii \undefined \def \showISBNxiii  #1{\unskip}     \fi
\ifx \showISSN     \undefined \def \showISSN      #1{\unskip}     \fi
\ifx \showLCCN     \undefined \def \showLCCN      #1{\unskip}     \fi
\ifx \shownote     \undefined \def \shownote      #1{#1}          \fi
\ifx \showarticletitle \undefined \def \showarticletitle #1{#1}   \fi
\ifx \showURL      \undefined \def \showURL       {\relax}        \fi
\providecommand\bibfield[2]{#2}
\providecommand\bibinfo[2]{#2}
\providecommand\natexlab[1]{#1}
\providecommand\showeprint[2][]{arXiv:#2}

\bibitem[\protect\citeauthoryear{Ali, Haque, Zheng, Washington, and
  Yildirimoglu}{Ali et~al\mbox{.}}{2019}]%
        {ali2019hazard}
\bibfield{author}{\bibinfo{person}{Yasir Ali}, \bibinfo{person}{Md~Mazharul
  Haque}, \bibinfo{person}{Zuduo Zheng}, \bibinfo{person}{Simon Washington},
  {and} \bibinfo{person}{Mehmet Yildirimoglu}.}
  \bibinfo{year}{2019}\natexlab{}.
\newblock \showarticletitle{A hazard-based duration model to quantify the
  impact of connected driving environment on safety during mandatory
  lane-changing}.
\newblock \bibinfo{journal}{\emph{Transportation research part C: emerging
  technologies}}  \bibinfo{volume}{106} (\bibinfo{year}{2019}),
  \bibinfo{pages}{113--131}.
\newblock


\bibitem[\protect\citeauthoryear{Aoki, Higuchi, and Altintas}{Aoki
  et~al\mbox{.}}{2020}]%
        {aoki2020cooperative}
\bibfield{author}{\bibinfo{person}{Shunsuke Aoki}, \bibinfo{person}{Takamasa
  Higuchi}, {and} \bibinfo{person}{Onur Altintas}.}
  \bibinfo{year}{2020}\natexlab{}.
\newblock \showarticletitle{Cooperative perception with deep reinforcement
  learning for connected vehicles}. In \bibinfo{booktitle}{\emph{2020 IEEE
  Intelligent Vehicles Symposium (IV)}}. IEEE, \bibinfo{pages}{328--334}.
\newblock


\bibitem[\protect\citeauthoryear{Cao, B{\i}y{\i}k, Wang, Raventos, Gaidon,
  Rosman, and Sadigh}{Cao et~al\mbox{.}}{2020}]%
        {cao2020reinforcement}
\bibfield{author}{\bibinfo{person}{Zhangjie Cao}, \bibinfo{person}{Erdem
  B{\i}y{\i}k}, \bibinfo{person}{Woodrow~Z Wang}, \bibinfo{person}{Allan
  Raventos}, \bibinfo{person}{Adrien Gaidon}, \bibinfo{person}{Guy Rosman},
  {and} \bibinfo{person}{Dorsa Sadigh}.} \bibinfo{year}{2020}\natexlab{}.
\newblock \showarticletitle{Reinforcement learning based control of imitative
  policies for near-accident driving}.
\newblock \bibinfo{journal}{\emph{arXiv preprint arXiv:2007.00178}}
  (\bibinfo{year}{2020}).
\newblock


\bibitem[\protect\citeauthoryear{Chen, Liu, Lin, Huang, and Zhu}{Chen
  et~al\mbox{.}}{2023}]%
        {Chen2023Mixed}
\bibfield{author}{\bibinfo{person}{Pin-Chun Chen}, \bibinfo{person}{Xiangguo
  Liu}, \bibinfo{person}{Chung-Wei Lin}, \bibinfo{person}{Chao Huang}, {and}
  \bibinfo{person}{Qi Zhu}.} \bibinfo{year}{2023}\natexlab{}.
\newblock \showarticletitle{Mixed-Traffic Intersection Management Utilizing
  Connected and Autonomous Vehicles as Traffic Regulators}. In
  \bibinfo{booktitle}{\emph{Proceedings of the 28th Asia and South Pacific
  Design Automation Conference}} (Tokyo, Japan) \emph{(\bibinfo{series}{ASPDAC
  '23})}. \bibinfo{publisher}{Association for Computing Machinery},
  \bibinfo{address}{New York, NY, USA}, \bibinfo{pages}{52–57}.
\newblock
\showISBNx{9781450397834}
\urldef\tempurl%
\url{https://doi.org/10.1145/3566097.3567849}
\showDOI{\tempurl}


\bibitem[\protect\citeauthoryear{Dong, Chen, Li, Du, Steinfeld, and Labi}{Dong
  et~al\mbox{.}}{2020}]%
        {dong2020facilitating}
\bibfield{author}{\bibinfo{person}{Jiqian Dong}, \bibinfo{person}{Sikai Chen},
  \bibinfo{person}{Yujie Li}, \bibinfo{person}{Runjia Du},
  \bibinfo{person}{Aaron Steinfeld}, {and} \bibinfo{person}{Samuel Labi}.}
  \bibinfo{year}{2020}\natexlab{}.
\newblock \showarticletitle{Facilitating connected autonomous vehicle
  operations using space-weighted information fusion and deep reinforcement
  learning based control}.
\newblock \bibinfo{journal}{\emph{arXiv preprint arXiv:2009.14665}}
  (\bibinfo{year}{2020}).
\newblock


\bibitem[\protect\citeauthoryear{Du, Chen, Li, Ha, Dong, Anastasopoulos, and
  Labi}{Du et~al\mbox{.}}{2021}]%
        {du2021cooperative}
\bibfield{author}{\bibinfo{person}{Runjia Du}, \bibinfo{person}{Sikai Chen},
  \bibinfo{person}{Yujie Li}, \bibinfo{person}{Paul Young~Joun Ha},
  \bibinfo{person}{Jiqian Dong}, \bibinfo{person}{Panagiotis~Ch
  Anastasopoulos}, {and} \bibinfo{person}{Samuel Labi}.}
  \bibinfo{year}{2021}\natexlab{}.
\newblock \showarticletitle{A cooperative crash avoidance framework for
  autonomous vehicle under collision-imminent situations in mixed traffic
  stream}. In \bibinfo{booktitle}{\emph{2021 IEEE International Intelligent
  Transportation Systems Conference (ITSC)}}. IEEE,
  \bibinfo{pages}{1997--2002}.
\newblock


\bibitem[\protect\citeauthoryear{Fisac, Bajcsy, Herbert, Fridovich-Keil, Wang,
  Tomlin, and Dragan}{Fisac et~al\mbox{.}}{2018}]%
        {fisac2018probabilistically}
\bibfield{author}{\bibinfo{person}{Jaime~F Fisac}, \bibinfo{person}{Andrea
  Bajcsy}, \bibinfo{person}{Sylvia~L Herbert}, \bibinfo{person}{David
  Fridovich-Keil}, \bibinfo{person}{Steven Wang}, \bibinfo{person}{Claire~J
  Tomlin}, {and} \bibinfo{person}{Anca~D Dragan}.}
  \bibinfo{year}{2018}\natexlab{}.
\newblock \showarticletitle{Probabilistically safe robot planning with
  confidence-based human predictions}.
\newblock \bibinfo{journal}{\emph{arXiv preprint arXiv:1806.00109}}
  (\bibinfo{year}{2018}).
\newblock


\bibitem[\protect\citeauthoryear{Fu, Li, Luan, Zhang, and Mao}{Fu
  et~al\mbox{.}}{2018}]%
        {fu2018infrastructure}
\bibfield{author}{\bibinfo{person}{Yuchuan Fu}, \bibinfo{person}{Changle Li},
  \bibinfo{person}{Tom~H Luan}, \bibinfo{person}{Yao Zhang}, {and}
  \bibinfo{person}{Guoqiang Mao}.} \bibinfo{year}{2018}\natexlab{}.
\newblock \showarticletitle{Infrastructure-cooperative algorithm for effective
  intersection collision avoidance}.
\newblock \bibinfo{journal}{\emph{Transportation research part C: emerging
  technologies}}  \bibinfo{volume}{89} (\bibinfo{year}{2018}),
  \bibinfo{pages}{188--204}.
\newblock


\bibitem[\protect\citeauthoryear{Garcia, Molina-Galan, Boban, Gozalvez,
  Coll-Perales, {\c{S}}ahin, and Kousaridas}{Garcia et~al\mbox{.}}{2021}]%
        {garcia2021tutorial}
\bibfield{author}{\bibinfo{person}{Mario H~Castaqeda Garcia},
  \bibinfo{person}{Alejandro Molina-Galan}, \bibinfo{person}{Mate Boban},
  \bibinfo{person}{Javier Gozalvez}, \bibinfo{person}{Baldomero Coll-Perales},
  \bibinfo{person}{Taylan {\c{S}}ahin}, {and} \bibinfo{person}{Apostolos
  Kousaridas}.} \bibinfo{year}{2021}\natexlab{}.
\newblock \showarticletitle{A tutorial on 5G NR V2X communications}.
\newblock \bibinfo{journal}{\emph{IEEE Communications Surveys \& Tutorials}}
  (\bibinfo{year}{2021}).
\newblock


\bibitem[\protect\citeauthoryear{Ha, Chen, Dong, Du, Li, and Labi}{Ha
  et~al\mbox{.}}{2020}]%
        {ha2020leveraging}
\bibfield{author}{\bibinfo{person}{Paul Young~Joun Ha}, \bibinfo{person}{Sikai
  Chen}, \bibinfo{person}{Jiqian Dong}, \bibinfo{person}{Runjia Du},
  \bibinfo{person}{Yujie Li}, {and} \bibinfo{person}{Samuel Labi}.}
  \bibinfo{year}{2020}\natexlab{}.
\newblock \showarticletitle{Leveraging the capabilities of connected and
  autonomous vehicles and multi-agent reinforcement learning to mitigate
  highway bottleneck congestion}.
\newblock \bibinfo{journal}{\emph{arXiv preprint arXiv:2010.05436}}
  (\bibinfo{year}{2020}).
\newblock


\bibitem[\protect\citeauthoryear{Han and Miao}{Han and Miao}{2020}]%
        {han2020behavior}
\bibfield{author}{\bibinfo{person}{Songyang Han} {and} \bibinfo{person}{Fei
  Miao}.} \bibinfo{year}{2020}\natexlab{}.
\newblock \showarticletitle{Behavior planning for connected autonomous vehicles
  using feedback deep reinforcement learning}.
\newblock \bibinfo{journal}{\emph{arXiv preprint arXiv:2003.04371}}
  (\bibinfo{year}{2020}).
\newblock


\bibitem[\protect\citeauthoryear{Jiao, Liu, Sato, Chen, and Zhu}{Jiao
  et~al\mbox{.}}{2022a}]%
        {jiao2022semi}
\bibfield{author}{\bibinfo{person}{Ruochen Jiao}, \bibinfo{person}{Xiangguo
  Liu}, \bibinfo{person}{Takami Sato}, \bibinfo{person}{Qi~Alfred Chen}, {and}
  \bibinfo{person}{Qi Zhu}.} \bibinfo{year}{2022}\natexlab{a}.
\newblock \showarticletitle{Semi-supervised semantics-guided adversarial
  training for trajectory prediction}.
\newblock \bibinfo{journal}{\emph{arXiv preprint arXiv:2205.14230}}
  (\bibinfo{year}{2022}).
\newblock


\bibitem[\protect\citeauthoryear{Jiao, Liu, Zheng, Liang, and Zhu}{Jiao
  et~al\mbox{.}}{2022b}]%
        {jiao2022tae}
\bibfield{author}{\bibinfo{person}{Ruochen Jiao}, \bibinfo{person}{Xiangguo
  Liu}, \bibinfo{person}{Bowen Zheng}, \bibinfo{person}{Dave Liang}, {and}
  \bibinfo{person}{Qi Zhu}.} \bibinfo{year}{2022}\natexlab{b}.
\newblock \showarticletitle{Tae: A semi-supervised controllable behavior-aware
  trajectory generator and predictor}. In \bibinfo{booktitle}{\emph{2022
  IEEE/RSJ International Conference on Intelligent Robots and Systems (IROS)}}.
  IEEE, \bibinfo{pages}{12534--12541}.
\newblock


\bibitem[\protect\citeauthoryear{Kenney}{Kenney}{2011}]%
        {kenney2011DSRC}
\bibfield{author}{\bibinfo{person}{John~B Kenney}.}
  \bibinfo{year}{2011}\natexlab{}.
\newblock \showarticletitle{Dedicated short-range communications ({DSRC})
  standards in the {United States}}.
\newblock \bibinfo{journal}{\emph{Proc. IEEE}} \bibinfo{volume}{99},
  \bibinfo{number}{7} (\bibinfo{year}{2011}), \bibinfo{pages}{1162--1182}.
\newblock


\bibitem[\protect\citeauthoryear{Khayatian, Mehrabian, and
  Shrivastava}{Khayatian et~al\mbox{.}}{2018}]%
        {khayatian2018rim}
\bibfield{author}{\bibinfo{person}{Mohammad Khayatian},
  \bibinfo{person}{Mohammadreza Mehrabian}, {and} \bibinfo{person}{Aviral
  Shrivastava}.} \bibinfo{year}{2018}\natexlab{}.
\newblock \showarticletitle{RIM: Robust intersection management for connected
  autonomous vehicles}. In \bibinfo{booktitle}{\emph{2018 IEEE Real-Time
  Systems Symposium (RTSS)}}. IEEE, \bibinfo{pages}{35--44}.
\newblock


\bibitem[\protect\citeauthoryear{Li, Sun, Zhan, and Tomizuka}{Li
  et~al\mbox{.}}{2020a}]%
        {li2020interaction}
\bibfield{author}{\bibinfo{person}{Jinning Li}, \bibinfo{person}{Liting Sun},
  \bibinfo{person}{Wei Zhan}, {and} \bibinfo{person}{Masayoshi Tomizuka}.}
  \bibinfo{year}{2020}\natexlab{a}.
\newblock \showarticletitle{Interaction-aware behavior planning for autonomous
  vehicles validated with real traffic data}. In
  \bibinfo{booktitle}{\emph{Dynamic Systems and Control Conference}},
  Vol.~\bibinfo{volume}{84287}. American Society of Mechanical Engineers,
  \bibinfo{pages}{V002T31A005}.
\newblock


\bibitem[\protect\citeauthoryear{Li, Zhan, Sun, Chan, and Tomizuka}{Li
  et~al\mbox{.}}{2020b}]%
        {li2020adaptive}
\bibfield{author}{\bibinfo{person}{Zhaoting Li}, \bibinfo{person}{Wei Zhan},
  \bibinfo{person}{Liting Sun}, \bibinfo{person}{Ching-Yao Chan}, {and}
  \bibinfo{person}{Masayoshi Tomizuka}.} \bibinfo{year}{2020}\natexlab{b}.
\newblock \showarticletitle{Adaptive sampling-based motion planning with a
  non-conservatively defensive strategy for autonomous driving}.
\newblock \bibinfo{journal}{\emph{IFAC-PapersOnLine}} \bibinfo{volume}{53},
  \bibinfo{number}{2} (\bibinfo{year}{2020}), \bibinfo{pages}{15632--15638}.
\newblock


\bibitem[\protect\citeauthoryear{Liu, Huang, Wang, Zheng, and Zhu}{Liu
  et~al\mbox{.}}{2022a}]%
        {liu2022physics}
\bibfield{author}{\bibinfo{person}{Xiangguo Liu}, \bibinfo{person}{Chao Huang},
  \bibinfo{person}{Yixuan Wang}, \bibinfo{person}{Bowen Zheng}, {and}
  \bibinfo{person}{Qi Zhu}.} \bibinfo{year}{2022}\natexlab{a}.
\newblock \showarticletitle{Physics-Aware Safety-Assured Design of Hierarchical
  Neural Network based Planner}.
\newblock \bibinfo{journal}{\emph{2022 ACM/IEEE 13th International Conference
  on Cyber-Physical Systems (ICCPS)}} (\bibinfo{year}{2022}),
  \bibinfo{pages}{137--146}.
\newblock


\bibitem[\protect\citeauthoryear{Liu, Jiao, Zheng, Liang, and Zhu}{Liu
  et~al\mbox{.}}{2023}]%
        {liu2022neural}
\bibfield{author}{\bibinfo{person}{Xiangguo Liu}, \bibinfo{person}{Ruochen
  Jiao}, \bibinfo{person}{Bowen Zheng}, \bibinfo{person}{Dave Liang}, {and}
  \bibinfo{person}{Qi Zhu}.} \bibinfo{year}{2023}\natexlab{}.
\newblock \showarticletitle{Safety-Driven Interactive Planning for Neural
  Network-Based Lane Changing}. In \bibinfo{booktitle}{\emph{Proceedings of the
  28th Asia and South Pacific Design Automation Conference}} (Tokyo, Japan)
  \emph{(\bibinfo{series}{ASPDAC '23})}. \bibinfo{publisher}{Association for
  Computing Machinery}, \bibinfo{address}{New York, NY, USA},
  \bibinfo{pages}{39–45}.
\newblock
\showISBNx{9781450397834}
\urldef\tempurl%
\url{https://doi.org/10.1145/3566097.3567847}
\showDOI{\tempurl}


\bibitem[\protect\citeauthoryear{Liu, Masoud, and Zhu}{Liu
  et~al\mbox{.}}{2020}]%
        {liu2020impact}
\bibfield{author}{\bibinfo{person}{Xiangguo Liu}, \bibinfo{person}{Neda
  Masoud}, {and} \bibinfo{person}{Qi Zhu}.} \bibinfo{year}{2020}\natexlab{}.
\newblock \showarticletitle{Impact of sharing driving attitude information: A
  quantitative study on lane changing}. In \bibinfo{booktitle}{\emph{2020 IEEE
  Intelligent Vehicles Symposium (IV)}}. IEEE, \bibinfo{pages}{1998--2005}.
\newblock


\bibitem[\protect\citeauthoryear{Liu, Masoud, Zhu, and Khojandi}{Liu
  et~al\mbox{.}}{2022b}]%
        {liu2022markov}
\bibfield{author}{\bibinfo{person}{Xiangguo Liu}, \bibinfo{person}{Neda
  Masoud}, \bibinfo{person}{Qi Zhu}, {and} \bibinfo{person}{Anahita Khojandi}.}
  \bibinfo{year}{2022}\natexlab{b}.
\newblock \showarticletitle{A markov decision process framework to incorporate
  network-level data in motion planning for connected and automated vehicles}.
\newblock \bibinfo{journal}{\emph{Transportation Research Part C: Emerging
  Technologies}}  \bibinfo{volume}{136} (\bibinfo{year}{2022}),
  \bibinfo{pages}{103550}.
\newblock


\bibitem[\protect\citeauthoryear{Liu, Zhao, Masoud, and Zhu}{Liu
  et~al\mbox{.}}{2021a}]%
        {liu2021trajectory}
\bibfield{author}{\bibinfo{person}{Xiangguo Liu}, \bibinfo{person}{Guangchen
  Zhao}, \bibinfo{person}{Neda Masoud}, {and} \bibinfo{person}{Qi Zhu}.}
  \bibinfo{year}{2021}\natexlab{a}.
\newblock \showarticletitle{Trajectory Planning for Connected and Automated
  Vehicles: Cruising, Lane Changing, and Platooning}.
\newblock \bibinfo{journal}{\emph{SAE International Journal of Connected and
  Automated Vehicles}} \bibinfo{volume}{4}, \bibinfo{number}{12-04-04-0025}
  (\bibinfo{year}{2021}), \bibinfo{pages}{315--333}.
\newblock


\bibitem[\protect\citeauthoryear{Liu, Zhou, Wang, and Peeta}{Liu
  et~al\mbox{.}}{2021b}]%
        {liu2021proactive}
\bibfield{author}{\bibinfo{person}{Yongyang Liu}, \bibinfo{person}{Anye Zhou},
  \bibinfo{person}{Yu Wang}, {and} \bibinfo{person}{Srinivas Peeta}.}
  \bibinfo{year}{2021}\natexlab{b}.
\newblock \showarticletitle{Proactive longitudinal control of connected and
  autonomous vehicles with lane-change assistance for human-driven vehicles}.
  In \bibinfo{booktitle}{\emph{2021 IEEE International Intelligent
  Transportation Systems Conference (ITSC)}}. IEEE, \bibinfo{pages}{776--781}.
\newblock


\bibitem[\protect\citeauthoryear{Mirchevska, Pek, Werling, Althoff, and
  Boedecker}{Mirchevska et~al\mbox{.}}{2018}]%
        {mirchevska2018high}
\bibfield{author}{\bibinfo{person}{Branka Mirchevska},
  \bibinfo{person}{Christian Pek}, \bibinfo{person}{Moritz Werling},
  \bibinfo{person}{Matthias Althoff}, {and} \bibinfo{person}{Joschka
  Boedecker}.} \bibinfo{year}{2018}\natexlab{}.
\newblock \showarticletitle{High-level decision making for safe and reasonable
  autonomous lane changing using reinforcement learning}. In
  \bibinfo{booktitle}{\emph{2018 21st International Conference on Intelligent
  Transportation Systems (ITSC)}}. IEEE, \bibinfo{pages}{2156--2162}.
\newblock


\bibitem[\protect\citeauthoryear{Nassef, Sequeira, Salam, and Mahmoodi}{Nassef
  et~al\mbox{.}}{2020}]%
        {nassef2020building}
\bibfield{author}{\bibinfo{person}{Omar Nassef}, \bibinfo{person}{Luis
  Sequeira}, \bibinfo{person}{Elias Salam}, {and} \bibinfo{person}{Toktam
  Mahmoodi}.} \bibinfo{year}{2020}\natexlab{}.
\newblock \showarticletitle{Building a Lane Merge Coordination for Connected
  Vehicles Using Deep Reinforcement Learning}.
\newblock \bibinfo{journal}{\emph{IEEE Internet of Things Journal}}
  (\bibinfo{year}{2020}).
\newblock


\bibitem[\protect\citeauthoryear{Orzechowski, Meyer, and Lauer}{Orzechowski
  et~al\mbox{.}}{2018}]%
        {orzechowski2018tackling}
\bibfield{author}{\bibinfo{person}{Piotr~F Orzechowski},
  \bibinfo{person}{Annika Meyer}, {and} \bibinfo{person}{Martin Lauer}.}
  \bibinfo{year}{2018}\natexlab{}.
\newblock \showarticletitle{Tackling occlusions \& limited sensor range with
  set-based safety verification}. In \bibinfo{booktitle}{\emph{2018 21st
  International Conference on Intelligent Transportation Systems (ITSC)}}.
  IEEE, \bibinfo{pages}{1729--1736}.
\newblock


\bibitem[\protect\citeauthoryear{Park, Abdel-Aty, Wu, and Mattei}{Park
  et~al\mbox{.}}{2018}]%
        {park2018enhancing}
\bibfield{author}{\bibinfo{person}{Juneyoung Park}, \bibinfo{person}{Mohamed
  Abdel-Aty}, \bibinfo{person}{Yina Wu}, {and} \bibinfo{person}{Ilaria
  Mattei}.} \bibinfo{year}{2018}\natexlab{}.
\newblock \showarticletitle{Enhancing in-vehicle driving assistance information
  under connected vehicle environment}.
\newblock \bibinfo{journal}{\emph{IEEE Transactions on Intelligent
  Transportation Systems}} \bibinfo{volume}{20}, \bibinfo{number}{9}
  (\bibinfo{year}{2018}), \bibinfo{pages}{3558--3567}.
\newblock


\bibitem[\protect\citeauthoryear{Pek, Manzinger, Koschi, and Althoff}{Pek
  et~al\mbox{.}}{2020}]%
        {pek2020using}
\bibfield{author}{\bibinfo{person}{Christian Pek}, \bibinfo{person}{Stefanie
  Manzinger}, \bibinfo{person}{Markus Koschi}, {and} \bibinfo{person}{Matthias
  Althoff}.} \bibinfo{year}{2020}\natexlab{}.
\newblock \showarticletitle{Using online verification to prevent autonomous
  vehicles from causing accidents}.
\newblock \bibinfo{journal}{\emph{Nature Machine Intelligence}}
  \bibinfo{volume}{2}, \bibinfo{number}{9} (\bibinfo{year}{2020}),
  \bibinfo{pages}{518--528}.
\newblock


\bibitem[\protect\citeauthoryear{Shalev-Shwartz, Shammah, and
  Shashua}{Shalev-Shwartz et~al\mbox{.}}{2017}]%
        {shalev2017formal}
\bibfield{author}{\bibinfo{person}{Shai Shalev-Shwartz},
  \bibinfo{person}{Shaked Shammah}, {and} \bibinfo{person}{Amnon Shashua}.}
  \bibinfo{year}{2017}\natexlab{}.
\newblock \showarticletitle{On a formal model of safe and scalable self-driving
  cars}.
\newblock \bibinfo{journal}{\emph{arXiv preprint arXiv:1708.06374}}
  (\bibinfo{year}{2017}).
\newblock


\bibitem[\protect\citeauthoryear{Sun, Zhan, Chan, and Tomizuka}{Sun
  et~al\mbox{.}}{2019}]%
        {sun2019behavior}
\bibfield{author}{\bibinfo{person}{Liting Sun}, \bibinfo{person}{Wei Zhan},
  \bibinfo{person}{Ching-Yao Chan}, {and} \bibinfo{person}{Masayoshi
  Tomizuka}.} \bibinfo{year}{2019}\natexlab{}.
\newblock \showarticletitle{Behavior planning of autonomous cars with social
  perception}. In \bibinfo{booktitle}{\emph{2019 IEEE Intelligent Vehicles
  Symposium (IV)}}. IEEE, \bibinfo{pages}{207--213}.
\newblock


\bibitem[\protect\citeauthoryear{Wang, Huang, Wang, Wang, and Zhu}{Wang
  et~al\mbox{.}}{2022}]%
        {wang2022design}
\bibfield{author}{\bibinfo{person}{Yixuan Wang}, \bibinfo{person}{Chao Huang},
  \bibinfo{person}{Zhaoran Wang}, \bibinfo{person}{Zhilu Wang}, {and}
  \bibinfo{person}{Qi Zhu}.} \bibinfo{year}{2022}\natexlab{}.
\newblock \showarticletitle{Design-while-verify: correct-by-construction
  control learning with verification in the loop}. In
  \bibinfo{booktitle}{\emph{Proceedings of the 59th ACM/IEEE Design Automation
  Conference}}. \bibinfo{pages}{925--930}.
\newblock


\bibitem[\protect\citeauthoryear{Wang, Huang, and Zhu}{Wang
  et~al\mbox{.}}{2020}]%
        {Wang2020Energy}
\bibfield{author}{\bibinfo{person}{Yixuan Wang}, \bibinfo{person}{Chao Huang},
  {and} \bibinfo{person}{Qi Zhu}.} \bibinfo{year}{2020}\natexlab{}.
\newblock \showarticletitle{Energy-Efficient Control Adaptation with Safety
  Guarantees for Learning-Enabled Cyber-Physical Systems}. In
  \bibinfo{booktitle}{\emph{Proceedings of the 39th International Conference on
  Computer-Aided Design}} (Virtual Event, USA) \emph{(\bibinfo{series}{ICCAD
  '20})}. \bibinfo{publisher}{Association for Computing Machinery},
  \bibinfo{address}{New York, NY, USA}, Article \bibinfo{articleno}{22},
  \bibinfo{numpages}{9}~pages.
\newblock
\showISBNx{9781450380263}
\urldef\tempurl%
\url{https://doi.org/10.1145/3400302.3415676}
\showDOI{\tempurl}


\bibitem[\protect\citeauthoryear{Wang, Wu, Boriboonsomsin, Barth, Han, Kim, and
  Tiwari}{Wang et~al\mbox{.}}{2019}]%
        {wang2019cooperative}
\bibfield{author}{\bibinfo{person}{Ziran Wang}, \bibinfo{person}{Guoyuan Wu},
  \bibinfo{person}{Kanok Boriboonsomsin}, \bibinfo{person}{Matthew~J Barth},
  \bibinfo{person}{Kyungtae Han}, \bibinfo{person}{BaekGyu Kim}, {and}
  \bibinfo{person}{Prashant Tiwari}.} \bibinfo{year}{2019}\natexlab{}.
\newblock \showarticletitle{Cooperative ramp merging system: Agent-based
  modeling and simulation using game engine}.
\newblock \bibinfo{journal}{\emph{SAE International Journal of Connected and
  Automated Vehicles}} \bibinfo{volume}{2}, \bibinfo{number}{2}
  (\bibinfo{year}{2019}).
\newblock


\bibitem[\protect\citeauthoryear{Wu, Xiao, Chen, Zhang, and Liu}{Wu
  et~al\mbox{.}}{2020}]%
        {Wu2020Amphibious}
\bibfield{author}{\bibinfo{person}{Yaqi Wu}, \bibinfo{person}{Anxing Xiao},
  \bibinfo{person}{Haoyao Chen}, \bibinfo{person}{Shiwu Zhang}, {and}
  \bibinfo{person}{Yunhui Liu}.} \bibinfo{year}{2020}\natexlab{}.
\newblock \showarticletitle{Amphibious Robot’s Trajectory Tracking with
  DNN-Based Nonlinear Model Predictive Control}. In
  \bibinfo{booktitle}{\emph{2020 IEEE/ASME International Conference on Advanced
  Intelligent Mechatronics (AIM)}}. \bibinfo{pages}{2019--2024}.
\newblock
\urldef\tempurl%
\url{https://doi.org/10.1109/AIM43001.2020.9159003}
\showDOI{\tempurl}


\bibitem[\protect\citeauthoryear{Xiao, Luan, Zhao, Hong, Zhao, Chen, Wang, and
  Meng}{Xiao et~al\mbox{.}}{2022}]%
        {xiao2022robotic}
\bibfield{author}{\bibinfo{person}{Anxing Xiao}, \bibinfo{person}{Hao Luan},
  \bibinfo{person}{Ziqi Zhao}, \bibinfo{person}{Yue Hong},
  \bibinfo{person}{Jieting Zhao}, \bibinfo{person}{Weinan Chen},
  \bibinfo{person}{Jiankun Wang}, {and} \bibinfo{person}{Max Q-H Meng}.}
  \bibinfo{year}{2022}\natexlab{}.
\newblock \showarticletitle{Robotic autonomous trolley collection with
  progressive perception and nonlinear model predictive control}. In
  \bibinfo{booktitle}{\emph{2022 International Conference on Robotics and
  Automation (ICRA)}}. IEEE, \bibinfo{pages}{4480--4486}.
\newblock


\bibitem[\protect\citeauthoryear{Xing, Lv, and Cao}{Xing et~al\mbox{.}}{2019}]%
        {xing2019personalized}
\bibfield{author}{\bibinfo{person}{Yang Xing}, \bibinfo{person}{Chen Lv}, {and}
  \bibinfo{person}{Dongpu Cao}.} \bibinfo{year}{2019}\natexlab{}.
\newblock \showarticletitle{Personalized vehicle trajectory prediction based on
  joint time-series modeling for connected vehicles}.
\newblock \bibinfo{journal}{\emph{IEEE Transactions on Vehicular Technology}}
  \bibinfo{volume}{69}, \bibinfo{number}{2} (\bibinfo{year}{2019}),
  \bibinfo{pages}{1341--1352}.
\newblock


\bibitem[\protect\citeauthoryear{Zhan, Liu, Chan, and Tomizuka}{Zhan
  et~al\mbox{.}}{2016}]%
        {zhan2016non}
\bibfield{author}{\bibinfo{person}{Wei Zhan}, \bibinfo{person}{Changliu Liu},
  \bibinfo{person}{Ching-Yao Chan}, {and} \bibinfo{person}{Masayoshi
  Tomizuka}.} \bibinfo{year}{2016}\natexlab{}.
\newblock \showarticletitle{A non-conservatively defensive strategy for urban
  autonomous driving}. In \bibinfo{booktitle}{\emph{2016 IEEE 19th
  International Conference on Intelligent Transportation Systems (ITSC)}}.
  IEEE, \bibinfo{pages}{459--464}.
\newblock


\bibitem[\protect\citeauthoryear{Zhang, Tian, and Duffy}{Zhang
  et~al\mbox{.}}{2022a}]%
        {zhang2022trust}
\bibfield{author}{\bibinfo{person}{Zhengming Zhang}, \bibinfo{person}{Renran
  Tian}, {and} \bibinfo{person}{Vincent~G Duffy}.}
  \bibinfo{year}{2022}\natexlab{a}.
\newblock \showarticletitle{Trust in Automated Vehicle: A Meta-Analysis}.
\newblock In \bibinfo{booktitle}{\emph{Human-Automation Interaction:
  Transportation}}. \bibinfo{publisher}{Springer}, \bibinfo{pages}{221--234}.
\newblock


\bibitem[\protect\citeauthoryear{Zhang, Tian, Sherony, Domeyer, and Ding}{Zhang
  et~al\mbox{.}}{2022b}]%
        {zhang2022attention}
\bibfield{author}{\bibinfo{person}{Zhengming Zhang}, \bibinfo{person}{Renran
  Tian}, \bibinfo{person}{Rini Sherony}, \bibinfo{person}{Joshua Domeyer},
  {and} \bibinfo{person}{Zhengming Ding}.} \bibinfo{year}{2022}\natexlab{b}.
\newblock \showarticletitle{Attention-Based Interrelation Modeling for
  Explainable Automated Driving}.
\newblock \bibinfo{journal}{\emph{IEEE Transactions on Intelligent Vehicles}}
  (\bibinfo{year}{2022}).
\newblock


\bibitem[\protect\citeauthoryear{Zheng, Lin, Shiraishi, and Zhu}{Zheng
  et~al\mbox{.}}{2019}]%
        {zheng2019design}
\bibfield{author}{\bibinfo{person}{Bowen Zheng}, \bibinfo{person}{Chung-Wei
  Lin}, \bibinfo{person}{Shinichi Shiraishi}, {and} \bibinfo{person}{Qi Zhu}.}
  \bibinfo{year}{2019}\natexlab{}.
\newblock \showarticletitle{Design and analysis of delay-tolerant intelligent
  intersection management}.
\newblock \bibinfo{journal}{\emph{ACM Transactions on Cyber-Physical Systems}}
  \bibinfo{volume}{4}, \bibinfo{number}{1} (\bibinfo{year}{2019}),
  \bibinfo{pages}{1--27}.
\newblock


\bibitem[\protect\citeauthoryear{Zhou, Chen, Yan, Li, Yin, and Ge}{Zhou
  et~al\mbox{.}}{2022}]%
        {zhou2022multi}
\bibfield{author}{\bibinfo{person}{Wei Zhou}, \bibinfo{person}{Dong Chen},
  \bibinfo{person}{Jun Yan}, \bibinfo{person}{Zhaojian Li},
  \bibinfo{person}{Huilin Yin}, {and} \bibinfo{person}{Wanchen Ge}.}
  \bibinfo{year}{2022}\natexlab{}.
\newblock \showarticletitle{Multi-agent reinforcement learning for cooperative
  lane changing of connected and autonomous vehicles in mixed traffic}.
\newblock \bibinfo{journal}{\emph{Autonomous Intelligent Systems}}
  \bibinfo{volume}{2}, \bibinfo{number}{1} (\bibinfo{year}{2022}),
  \bibinfo{pages}{1--11}.
\newblock


\bibitem[\protect\citeauthoryear{Zhu, Huang, Jiao, Lan, Liang, Liu, Wang, Wang,
  and Xu}{Zhu et~al\mbox{.}}{2021}]%
        {zhu2021safety}
\bibfield{author}{\bibinfo{person}{Qi Zhu}, \bibinfo{person}{Chao Huang},
  \bibinfo{person}{Ruochen Jiao}, \bibinfo{person}{Shuyue Lan},
  \bibinfo{person}{Hengyi Liang}, \bibinfo{person}{Xiangguo Liu},
  \bibinfo{person}{Yixuan Wang}, \bibinfo{person}{Zhilu Wang}, {and}
  \bibinfo{person}{Shichao Xu}.} \bibinfo{year}{2021}\natexlab{}.
\newblock \showarticletitle{Safety-Assured Design and Adaptation of
  Learning-Enabled Autonomous Systems}. In \bibinfo{booktitle}{\emph{26th
  ACM/IEEE Asia and South Pacific Design Automation Conference (ASP-DAC)}}.
\newblock


\end{thebibliography}


\end{document}